\documentclass[times,review,10pt]{elsarticle}
\usepackage{amssymb}
\usepackage{amsmath}
\usepackage{graphicx}
\usepackage{booktabs}
\usepackage{algpseudocode}
\usepackage{amssymb}
\usepackage{amsmath}
\usepackage{bm}
\usepackage{multirow}
\usepackage{makecell}
\usepackage{hyperref}
\usepackage{pifont}
\usepackage{subfigure}
\usepackage{color}
\usepackage{wrapfig}
\usepackage{url}
\usepackage{hyperref}
\usepackage{color}

\newcommand{\blue}[1] {\textcolor{black}{{#1}}}
\usepackage[ruled,linesnumbered]{algorithm2e}
\usepackage[accsupp]{axessibility}
\journal{Pattern Recognition}
\begin{document}
\begin{frontmatter}
\title{Attacking Hard-Label Large Vision-Language Models with Model-Sensitive Adversarial Patch Designs}

\author[label1]{Nian Ai\corref{cor1}}
\author[label2]{Guangke Chen\corref{cor1}}
\author[label1]{Xiaowen Cai}
\author[label3]{Zhongliang Guo\corref{cor2}}
\author[label2]{Daizong Liu\corref{cor2}}
\author[label1]{Pan Zhou}
\author[label4]{Ognjen Arandjelovi\'c}

\affiliation[label1]{organization={School of Cyber Science and Engineering, Huazhong University of Science and Technology},
            city={Wuhan},
            postcode={430074},
            country={China}}
\affiliation[label2]{organization={Institute for Math \& AI, Wuhan University},
            city={Wuhan},
            postcode={430072},
            country={China}}
\affiliation[label3]{organization={School of Natural and Computing Sciences, University of Aberdeen},
            city={Aberdeen},
            country={United Kingdom}}
\affiliation[label4]{organization={School of Computer Science, University of St Andrews},
            city={St Andrews},
            country={United Kingdom}}

\cortext[cor1]{Equal Contributions.}
\cortext[cor2]{Corresponding Authors: Zhongliang Guo and Daizong Liu.}



\begin{abstract}
Large vision-language models (LVLMs) have demonstrated impressive capabilities in handling multi-modal downstream tasks, gaining increasing popularity.
However, recent studies show that LVLMs are susceptible to both intentional and inadvertent attacks.
Existing attackers ideally optimize adversarial perturbations with backpropagated gradients from LVLMs, thus limiting their scalability in practical scenarios as real-world LVLM applications will not provide any LVLM's gradient or details. 
Motivated by this research gap and counter-practical phenomenon, we propose a novel hard-label attack method for LVLMs, named \textit{HardPatch}, to generate visual adversarial patches by solely querying the model.
Our method provides deeper insights into how to investigate the vulnerability of LVLMs in local visual regions and generate corresponding adversarial substitution under the practical yet challenging hard-label setting. 
Specifically, we first split each image into uniform patches and mask each of them to individually assess their sensitivity to the LVLM model.
Then, according to the descending order of sensitive scores, we iteratively select the most vulnerable patch to initialize noise and estimate gradients with further additive random noises for optimization. 
In this manner, multiple patches are perturbed until the altered image satisfies the adversarial condition.
Extensive LVLM models and datasets are evaluated to demonstrate the adversarial nature of \textit{HardPatch}.
Our empirical observations suggest that with appropriate patch substitution and optimization, \textit{HardPatch} can craft effective adversarial images to attack hard-label LVLMs.
\end{abstract}





\end{frontmatter}



\section{Introduction}
Nowadays, large vision-language models (LVLMs) \cite{dai2024instructblip,liu2024visual,zhu2023minigpt,fu2026hyperr3snet,zuo2026autonomous,liu2021spatiotemporal,liu2021context,liu2020jointly}, at the juncture of computer vision and natural language processing, 
have become indispensable and marked a significant milestone in the field of artificial intelligence.
By further benefiting from the strong comprehension of large language models (LLMs) \cite{brown2020language,gao6045154multi,gao2025distributed,chen2025learning}, recent LVLMs \cite{dai2024instructblip,liu2024visual,zhu2023minigpt} on top of LLMs show notable developments in numerous downstream tasks \cite{liu2026spatial,ramesh2022hierarchical,li2023blip,yang2025reid,liu2025underwater,liu2024sentiment}.
However, most recently proposed LVLMs suffer from severe security issues \cite{liu2024safety,fan2024unbridled,liu2026generating,liu2024explicitly,liu2023robust,tao20233dhacker,hu2022exploring,liu2023point,liu2022imperceptible,liu2025seeing}, where an attacker's well-crafted adversarial input sample can easily fool the LVLM models, posing a considerable challenge to real-world LVLM applications.

\begin{figure}
\centering
\includegraphics[width=0.66\columnwidth]{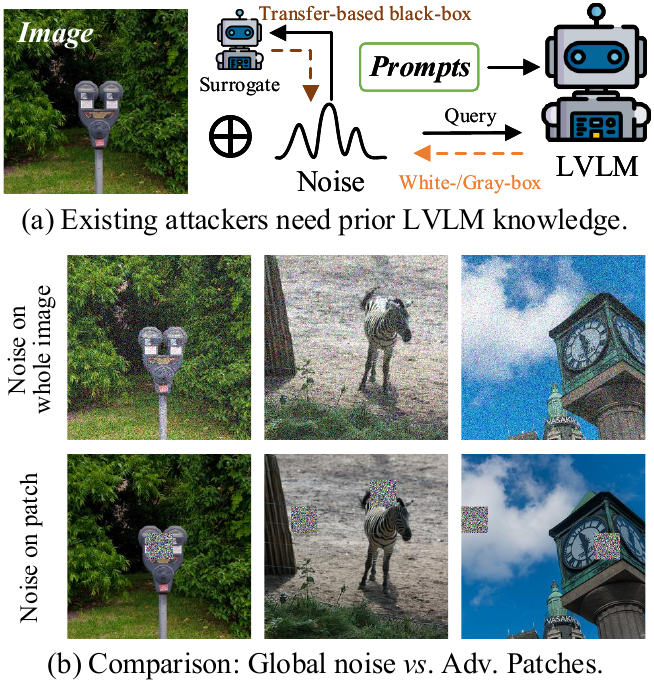}
\vspace{-10pt}
\caption{(a) Attack process of existing LVLM attackers. (b) We re-implement \cite{zhao2024evaluating} in the hard-label setting by removing its surrogate model. Compared to it, our adversarial patches have fewer perturbations and are easier to add.}
\vspace{-10pt}
    \label{fig:intro}
\end{figure}

Based on the accessibility level of victim models, existing LVLM attackers can be generally categorized into three types: white-box attacks \cite{bailey2023image,dong2023robust,liu2026large,cui2023robustness,gao2024adversarial,wang2024stop,lu2024test,luo2024image,gao2024inducing}, gray-box attacks \cite{shayegani2023jailbreak,yan2025fit}, and transfer-based black-box attacks \cite{zhao2024evaluating,yin2023vlattack}, as shown in Figure~\ref{fig:intro} (a).
For white-box attacks, the attackers are assumed to have full knowledge of the victim LVLMs, including model architecture and parameters. These works formulate the attack as an optimization problem and utilize the backpropagated gradient to generate adversarial examples. 
To alleviate this reliance on model details to a certain extent, gray-box attacks \cite{shayegani2023jailbreak,yan2025fit} solely require access to the visual encoder of LVLMs.
However, since real-world LVLM applications are impossible to share any model details with users, white-/gray-box attacks seem excessively idealistic and cannot work well in practical scenarios.
Although no target-model details are required in transfer-based black-box attacks \cite{zhao2024evaluating,dong2023robust}, they still rely on the additional knowledge of other surrogate LVLM models.
In sum, existing LVLM attackers are severely limited by their scalability, and there is no attack that truly does not require any prior LVLM information in a more challenging hard-label setting \cite{cheng2018query}.

To address this research gap, we introduce the first hard-label adversarial attack against LVLMs, where the attackers can solely query the input/output of LVLMs.
However, without using model details, it is difficult to determine where and how to add perturbations to images to mislead LVLMs.
Luckily, the design of adversarial patch provides a concise and interpretable way to achieve successful real-world attacks \cite{brown2017adversarial,duan2020adversarial}.
By appropriately placing the adversarial patches on the image according to the model's attention, its adversarial nature will fool the LVLM's eyes and lead to inaccurate prompt reasoning.
Moreover, we empirically find that adversarial patches have fewer perturbations and are easier to add than directly perturbing pixel-wise noises on whole images \cite{zhao2024evaluating,cheng2018query}, as shown in Figure~\ref{fig:intro} (b).
Based on the above observations, we attempt to investigate \textit{``How to design effective adversarial patches to mislead hard-label LVLMs?"}.
Therefore, the remaining questions in designing LVLM attacks are: In the hard-label setting, (1) how to explore the LVLM's attention on different local regions of images for patch substitution?
and (2) how to design/optimize the patch pattern in order to achieve the adversarial condition?

In this paper, we propose a novel adversarial patch method called \textit{HardPatch} to tackle the above hard-label issues.
Specifically, we first uniformly split the input image into multiple patches with the same size.
Then, to assess the sensitivity of each patch to the LVLM model, we individually mask each patch and feed them into the LVLM to measure the semantic changes between their corresponding text output and the original output.
The larger the distance, the more sensitive the LVLM model is to altering the corresponding patch.
Therefore, by scoring all patches according to their sensitivities in descending order, we iteratively substitute the more vulnerable patch with initial noise and estimate gradients with further additive random noises for optimizing the adversarial pattern. If the patch updated with a fixed number of iterations is still not adversarial, we additionally perform the same altering process on the next patch. Multiple patches are perturbed until the altered image satisfies the adversarial condition. 
\blue{Unlike conventional vision attacks that rely on classification confidence or visual features, our method explicitly leverages the semantic sensitivity of generated text outputs, thereby targeting the unique cross-modal alignment mechanism in LVLMs.}
The key contributions of our work are outlined as follows: 
\begin{itemize}
    \item  We design \textit{HardPatch}, a novel adversarial attack method for more practical yet challenging hard-label LVLMs. We propose to generate visual adversarial patches to be added to input images for attackers in real-world scenarios.
    \item To determine where to place the adversarial patch, we develop a replacement order determination module to investigate the sensitivity of LVLM to each patch. Based on this, we iteratively substitute more vulnerable patches with noise and design the gradient estimation strategy to further optimize it until the attack succeeds.
    \item These insights are validated by extensive experiments on different LVLM models and datasets. Corresponding results demonstrate the effectiveness of our proposed \textit{HardPatch} against hard-label LVLM models.
\end{itemize}

\section{Related Work}
\noindent \textbf{Adversarial Robustness of LVLM Models.}
LVLMs generally combine the capabilities of processing visual information with natural language understanding by using pre-trained vision encoders with language models.
Due to this multimodal nature \cite{szegedy2013intriguing}, LVLMs are particularly vulnerable as the multi-modal integration not only amplifies their vulnerable utility but also introduces new attack vectors that are absent in unimodal systems.
Most of existing LVLM attackers \cite{bailey2023image,dong2023robust,liu2026large,cui2023robustness,gao2024adversarial,wang2024stop,lu2024test,luo2024image,gao2024inducing,liu2026attacking2,liu2026survey,liu2026flexible,liu2026understanding,liu2026forbid,liu2026attacking,cai2025towards,liu2025survey,liu2024pandora,yam2025myface,guo2026artwork,guo2024laaca,liu2026diffprotect,guo2025pca,lin2026priorasskill,qian2026tdcount,wang2026harmreduction,zhao2026exploring,zhao2026metacognitive,yang2026pointcloud,zhou2026posecompass,li2026fedkd,zhao2026p2p,duan2026gaitprotector,lei2026gazesync,li2026safetwinnet,li2026w2sdefense,guo2025thesis,hu2025syntactic,li2025guardfed,guo2025beyond,li2025fairseg,zhao2025roptester,lei2025transferable,li2025flsurvey,qian2025perspective,qian2025t2icount,lei2025instant,zhao2025backdoorsurvey,liu2024mmadpurify,zhao2024weaktostrong,guo2024signature,qian2024semisupervised,guo2023siamese} are inspired by the adversarial vulnerability observed in vision tasks.
They evaluate the adversarial robustness of LVLMs under white-box settings, where they have the full knowledge of LVLMs models including network structure and weights. To generate the adversarial examples, they simply add and optimize imperceptible perturbations on the whole image to benign image inputs via back-propagation.
To reduce the reliance on model knowledge, some gray-box attackers \cite{shayegani2023jailbreak,yan2025fit} solely require access to the visual encoder of LVLMs and directly generate the perturbed visual representations to fool the latter process.
Although a few researchers \cite{zhao2024evaluating,yin2023vlattack} claim that they achieve more challenging black-box attacks, their attacks are implemented in a transfer-based setting, where they still require the additional knowledge of other surrogate LVLM models to generate adversarial samples then transfer them to attack victim LVLMs. 

\noindent \textbf{Adversarial Patch.}
Adversarial patches \cite{brown2017adversarial,karmon2018lavan,eykholt2018robust} represent a unique subclass of adversarial attacks that focus on generating localized perturbations to fool deep learning models. Unlike traditional adversarial attacks, which often involve slight pixel-level modifications across the entire image, adversarial patches are confined to small regions but can cause significant misclassifications even when covering only a fraction of the input.
This adversarial patch is proven to have more practicality \cite{athalye2018synthesizing}, contributing to a deeper understanding of the interaction between digital perturbations and physical environments. Some works \cite{liu2016delving} also explore the transferability of adversarial patches across different models.
Concurrently, \cite{duan2020adversarial} focused on generating adversarial patches using generative models, enhancing the efficiency and effectiveness of attack generation.
However, there is still no adversarial patch attack being investigated in LVLMs.

\section{The Proposed Attack}

\subsection{Preliminary}
Given the input image $\bm{x}$ and the input prompt $\bm{c}_{in}$, an image-grounded text generative LVLM $f_{\Theta}(\bm{x},\bm{c}_{in}) \mapsto \bm{c}_{out}$ predicts a suitable textual response $\bm{c}_{out}$, where $\Theta$ is the LVLM's parameters. Since LVLM drivers multiple tasks, in image captioning tasks, for instance, $\bm{c}_{in}$ is a placeholder $\oslash$ and $\bm{c}_{out}$ is the caption; in visual question answering tasks, $\bm{c}_{in}$ is the question and $\bm{c}_{out}$ is the answer.
The adversary typically adds an imperceptible visual perturbation on the benign image to craft an adversarial example $\bm{x}'$ that misleads the LVLM model $f_{\Theta}$ to output a wrong prediction with a specific prompt $\bm{c}_{in}$ as:
\begin{equation}
f_{\Theta}(\bm{x}',\bm{c}_{in}) \neq f_{\Theta}(\bm{x},\bm{c}_{in}), \ \text{s.t.} \ \bm{x}' = \bm{x} + \bm{\delta},
\end{equation}
where $ \bm{\delta}$ is the visual perturbation (\textit{e.g.}, global noise or local patch).
Specifically, for the untargeted attack, the attack is successful if the model is misled to generate text different from the prediction with the clean image. For the targeted attack, the attack is considered to be successful only if the prediction exactly matches the attackers' preset target text $\bm{c}_{out}'$ where $\bm{c}_{out}' \neq \bm{c}_{out}$.

\begin{figure}[t!]
\centering
\includegraphics[width=0.99\textwidth]{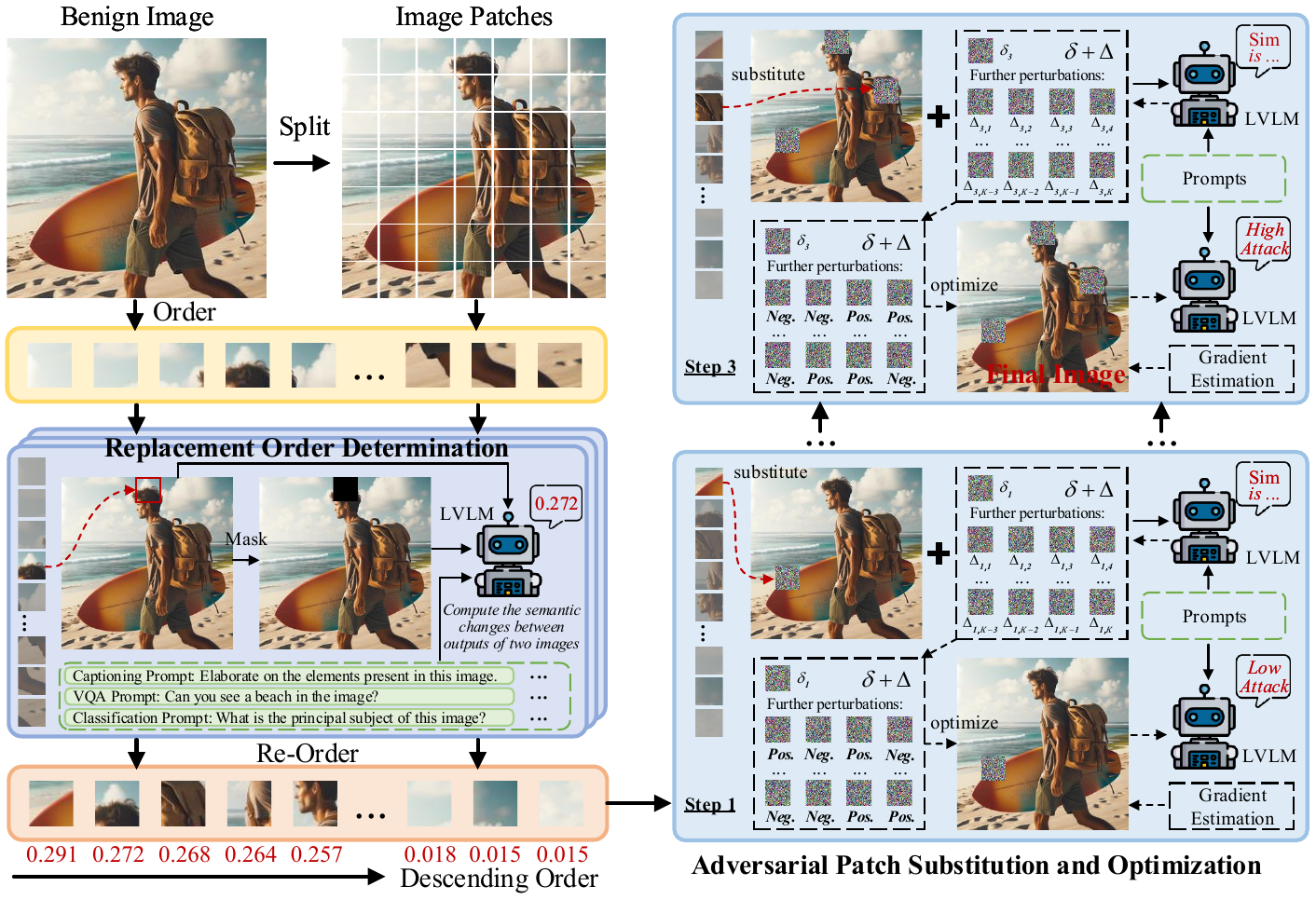}
\caption{Overview of our proposed \textit{HardPatch} attack. Given the input image and prompts, we first uniformly split the image into patches of the same size. Then, we individually mask each patch to assess their sensitivity to the LVLM model by measuring the semantic changes between their text output with the clean one. After that, we iteratively substitute the most vulnerable patch with noise and estimate gradients to update its noisy pattern. Multiple patches are perturbed until the final altered image achieves the adversarial condition.}
\vspace{-6pt}
    \label{fig:pipeline}
\end{figure} 

In this paper, we focus on the task of hard-label LVLM adversarial attack, \textit{i.e.}, attackers can only access to the predicted text output from the victim LVLM model to generate adversarial examples.

\subsection{Overview of Our \textit{HardPatch} Attack}
\noindent \textbf{Discussion on Our Motivation.}
Existing LVLM attackers \cite{dong2023robust,yan2025fit,wang2024stop,zhang2024b,luo2024image,zhao2024evaluating} generally add pixel-wise noise on the whole image input, which are easily optimized in the white-/gray-box or transfer-based black-box setting via the backpropagated gradient. However, in more challenging hard-label setting, it is difficult to directly determine and tamper the LVLM's adversarial attention to optimize previous global noise by solely querying the LVLM model. 
Inspired by the global semantic invariant characteristic with local contexts mask of MAE \cite{he2022masked}, we propose to develop attack based on adversarial patch, which assesses the LVLM's vulnerability on local alteration by individually masking different patches of the original images. Then, the patches that have a greater adversarial impact on the LVLM model will be further combined to jointly be perturbed for attacks.
\blue{We note that our patch sensitivity estimation evaluates each region independently, which does not explicitly capture higher-order interactions among patches. While such interactions may exist, modeling them directly would require evaluating combinations of patches, leading to a combinatorial increase in query cost that is infeasible under the hard-label setting.
Instead, we adopt an efficient first-order approximation by ranking patches based on their individual impact on the model output. Importantly, the subsequent optimization stage jointly updates perturbations over the selected patches, during which interactions are implicitly captured through iterative feedback from the model.}

\noindent \textbf{Overall of Our Attack Pipeline.}
The overall pipeline of our \textit{HardPatch} is illustrated in Figure~\ref{fig:pipeline}. A placement order determination module is first introduced to assess the sensitivity of each patch to the LVLM and re-order the patches. Then, the adversarial patch substitution and optimization module is proposed to alter the patches following the order step-by-step. Multiple patches are perturbed until the attack succeeds.
We will provide more details of these two modules in the following.

\subsection{Replacement Order Determination of Adversarial Patches}
As for initialization, we first uniformly split the image $\bm{x}$ into $M$ patches $\{\bm{v}_1,\bm{v}_2,...,\bm{v}_M\}$.
\blue{We partition images by a fixed number of patches rather than fixed patch size, making the method naturally adaptive to different image resolutions.}
Then, we propose to individually mask each original patch to assess the impact of the corresponding altered sample, where the larger the impact, the more sensitive the LVLM model is to altering the corresponding patch.
Therefore, more important patches with greater impact on the victim model should be substituted with noisy patterns at the beginning in the adversarial replacement order.
Specifically, to evaluate the importance/sensitivity of each patch $\bm{v}_m,m \in M$, we set the patch $\bm{v}_m$ to be all zero and feed the image into the LVLM model.
\blue{We set the masked patch to zero values to remove its visual information without introducing additional semantic artifacts, enabling a more faithful estimation of its contribution to the generated output.}
We utilize a lightweight textual encoder (\textit{i.e.}, CLIP \cite{radford2021learning}) to evaluate the semantic similarity between its text output and the clean output as:
\begin{equation}
\label{eq:1}
    \mathcal{S}(\bm{v}_m) = \text{Sim}(f_{\Theta}(\bm{x}'(\bm{v}_m),\bm{c}_{in}),f_{\Theta}(\bm{x},\bm{c}_{in})),
\end{equation}
where $\bm{x}'(\bm{v}_m)$ denotes generating adversarial sample by altering patch $\bm{v}_m$, $\text{Sim}(\cdot)$ is the text-aware cosine similarity function and its range is between $[0,1]$. 
\blue{
Different from traditional vision models where importance is measured via logits or gradients, our sensitivity metric is derived from semantic changes in generated text, which inherently reflects the cross-modal reasoning process in LVLMs. This design enables us to capture LVLM-specific vulnerabilities without accessing internal representations.
}
Then we compute the importance score of each $\bm{v}_m$ by evaluating the semantic changes by altering patch $\bm{v}_m$, the large score indicates the better attack performance:
\begin{equation}
\label{eq:2}
    \mathcal{I}(\bm{v}_m) = 1 - \mathcal{S}(\bm{v}_m).
\end{equation}
Based on all importance scores $\{\mathcal{I}(\bm{v}_m)\}_{m=1}^{m=M}$, we sort all patches in descending order as the adversarial replacement order $\mathcal{O}=\{\bm{v}_1',\bm{v}_2',...,\bm{v}_M'\}$ for latter process.

\subsection{Adversarial Patch Substitution and Optimization}
To achieve the challenging hard-label LVLM attack, according to the replacement order $\mathcal{O}=\{\bm{v}_1',\bm{v}_2',...,\bm{v}_M'\}$, we propose to constantly substitute and optimize the most vulnerable patches to query the model for investigating whether the alter can change the output semantics.
Beginning at the first patch $\bm{v}_1'$, we first randomly sample patch-wise noise $\bm{\delta}_1$ from a uniform distribution to substitute $\bm{v}_1'$ in the image $\bm{x}$, then conduct $T$-step gradient estimation to update $\bm{\delta}_1$ by solely querying the LVLM model. If the $T$-times updated $\bm{\delta}_1$ can not achieve significant attack performance, we additionally substitute and optimize the latter patch with the same process.
The whole attacking procedure of adversarial patch substitution and optimization does not end until the adversarial condition is achieved.

\begin{algorithm}[t!]
    \SetAlgoLined
    \caption{Algorithm of The Proposed Attack } \label{alg1}
    \LinesNumbered
    \KwIn{Image input $\bm{x}$, text input $\bm{c}_{in}$, LVLM model $f_{\Theta}(\cdot)$}
    \KwOut{Adversarial image with perturbed patches}
    Split image $\bm{x}$ into $M$ Patches;
    
    \For(\tcp*[f]{Replacement Order Determination}){each patch $\bm{v}_m$ in $\bm{x}$}{ 
		Compute the importance score $\mathcal{I}(\bm{v}_m)$ via Eq.~(\ref{eq:1}),(\ref{eq:2});
    }
    Sort all patches based on their importance scores in descending order;
    
    \For(\tcp*[f]{Adversarial Patch Substitution and Optimization}){each patch in replacement order}{ 
    Replace patch $\bm{v}_m'$ with initial noise $\bm{\delta}_m$ on the image $\bm{x}_m$;
    
		\For{$t=1:T$}{
                Optimize $\bm{\delta}_m$ with a set of slight perturbations $\{\bm{\Delta}_k\}_{k=1}^{k=K}$ via Eq.~(\ref{eq:3}),(\ref{eq:4}),(\ref{eq:5});
        }
        \If{adversarial condition is satisfied (i.e., $\text{Sim}^{Tar} > \tau_1 \ or \ \text{Sim}^{Untar} < \tau_2$) \ or \  Adversarial patch number reaches preset \textit{Maximum}}  {
        break;
        }
            
    }
\Return The final $\bm{x}_m'$ is the adversarial sample
\end{algorithm}

In particular, as for the $m$-th order patch $\bm{v}_m'$, patch-wise noise $\bm{\delta}_m$ is initialized to substitute $\bm{v}_m'$ and we can further optimize it with a reasonable direction by querying the LVLM with additive random noise. Specifically, we first employ a normalized uniform distribution $\bm{u}\cdot \exp(\bm{u}-1),\bm{u}\sim \mathcal{U}(-1,1)$ to add a set of slight perturbations $\{\bm{\Delta}_k\}_{k=1}^{k=K}$ on the patch $\bm{\delta}_m$ for further altering. 
\blue{We initialize the adversarial patch with a uniform distribution to provide an unbiased starting point, allowing the optimization to explore the perturbation space without prior structural constraints.}
At the $t$-th step, we define an indicator function $\varphi_k$ to measure whether the perturbation $\bm{\Delta}_k$ can cause the misprediction of LVLM model as:
\begin{equation}
\label{eq:3}
 \varphi^{Tar}_k=
    \left\{\begin{array}{ll} 
        1, & \text{If} \ \text{Sim}(f_{\Theta}(\bm{x}_{m}'(\bm{\delta}_m+\bm{\Delta}_k)),\bm{c}_{out}')>\text{Sim}(f_{\Theta}(\bm{x}_{m}'(\bm{\delta}_m)),\bm{c}_{out}'),
         \\ 0, & \text{If} \ \text{Sim}(f_{\Theta}(\bm{x}_{m}'(\bm{\delta}_m+\bm{\Delta}_k)),\bm{c}_{out}')\leq \text{Sim}(f_{\Theta}(\bm{x}_{m}'(\bm{\delta}_m)),\bm{c}_{out}'),
    \end{array}\right.
\end{equation}
\begin{equation}
\label{eq:4}
  \varphi^{Untar}_k=
    \left\{\begin{array}{ll} 
        1, & \text{If} \ \text{Sim}(f_{\Theta}(\bm{x}_{m}'(\bm{\delta}_m+\bm{\Delta}_k)),\bm{c}_{out})<\text{Sim}(f_{\Theta}(\bm{x}_{m}'(\bm{\delta}_m)),\bm{c}_{out}),
         \\ 0, & \text{If} \ \text{Sim}(f_{\Theta}(\bm{x}_{m}'(\bm{\delta}_m+\bm{\Delta}_k)),\bm{c}_{out})\geq \text{Sim}(f_{\Theta}(\bm{x}_{m}'(\bm{\delta}_m)),\bm{c}_{out}),
    \end{array}\right.
\end{equation}
where $Tar,Untar$ denote the targeted and untargeted attacks, $\bm{x}_{m}'$ denotes the image being substituted by previous patch-wise perturbations with $\{\bm{\delta}_1+\bm{\Delta},\bm{\delta}_2+\bm{\Delta},...,\bm{\delta}_{m-1}+\bm{\Delta}\}$. Therefore, following the traditional Monte Carlo method \cite{james1980monte}, we estimate the final updating
direction by weighted averaging over the $K$ possible directions $\{\bm{\Delta}_k\}_{k=1}^{k=K}$, and optimize $\bm{\delta}_m$ as:
\begin{equation}
\label{eq:5}
    \bm{\delta}_m' = \bm{\delta}_m +\frac{\frac{1}{K}\sum_{k=1}^{K}\varphi_k\bm{\Delta}_k}{||\frac{1}{K}\sum_{k=1}^{K}\varphi_k\bm{\Delta}_k||_2}.
\end{equation}
By iteratively substituting and optimizing each patch with a set of perturbations with $T$-step, we can generate harmful noise with a certain number of perturbed patches to mislead the LVLM model. 
The overall algorithm of our attack is summarized in Algorithm~\ref{alg1}.

\subsection{\blue{Discussion on the Design Choice}}
\blue{
We note that some components in our framework, such as region sensitivity analysis and gradient estimation, have been explored in prior adversarial attack literature. However, our formulation is fundamentally different due to the hard-label LVLM setting, where standard signals such as logits, confidence scores, or attention maps are unavailable.
To address this, we redefine region sensitivity based on \emph{semantic changes in generated textual outputs} after masking image patches, shifting from visual/logit-level importance to language-conditioned semantic relevance. Similarly, instead of relying on gradient signals derived from model outputs, we construct a \emph{language-guided optimization objective} based on semantic similarity between generated responses, enabling effective optimization under hard-label constraints.
Importantly, these two components are jointly designed: region sensitivity reduces the search space over image patches, while language-guided feedback enables meaningful perturbation updates. This coupling is essential for efficient attack in LVLMs.
Overall, our method is not a direct reuse of existing techniques, but a reformulation of adversarial attack strategies tailored to multimodal hard-label settings, with redesigned sensitivity estimation and optimization objectives aligned with LVLM characteristics.}

\subsection{\blue{Discussion on Perceptibility}}
\blue{While patch-based perturbations may appear locally noticeable, perceptibility is jointly determined by both perturbation intensity and spatial extent. Under the same perturbation budget, global-noise-based methods typically modify nearly all pixels, leading to accumulated visual artifacts across the entire image. In contrast, our method perturbs only a small number of localized patches, leaving most regions unchanged and thus constraining the overall visual disruption. 
Moreover, in the hard-label setting, global noise often requires larger magnitude or more extensive perturbations to achieve comparable attack success, which further increases its visibility. Therefore, although patch perturbations are locally salient, global perturbations tend to introduce more widespread and noticeable artifacts under the same attack constraints.}

\section{Experiments}
\subsection{Experimental Setups}
\noindent \textbf{LVLM Models and Datasets.}
To assess the LVLMs' robustness against our attack, We consider four open-source and advanced LVLM models as
our evaluation benchmark, including BLIP-2 \cite{li2023blip} (OPT-2.7B), MiniGPT-4  \cite{zhu2023minigpt} (LLama 2-7B-Chat), LLaVA-1.5 \cite{liu2024visual} (Vicuna-7B), and InstructBLIP \cite{dai2024instructblip} (Vicuna-7B). 
As for LVLM datasets, we consider three datasets, \textit{i.e.}, MS-COCO \cite{lin2014microsoft}, ImageNet \cite{deng2009imagenet}, and DALL-E \cite{ramesh2021zero,ramesh2022hierarchical} with tasks of image classification, image captioning, and visual question answering (VQA). Specifically, We follow previous work \cite{luo2024image} and \cite{zhao2024evaluating} to construct MS-COCO and ImageNet datasets, respectively.
The DALL-E dataset employs a generative method, using random textual descriptions extracted from MS-COCO captions as prompts for image generation powered by GPT-4 \cite{achiam2023gpt}. Additionally, it includes randomly generated QA pairs based on the images.

\begin{table}[t!]
\small
\centering
\caption{Attack performance on different LVLM models. As for targeted attack $(\uparrow)$, we report the semantic similarity scores between the LVLM's output and the attackers' chosen label ``Unknown". As for untargeted attack $(\downarrow)$, we report the semantic similarity scores between the LVLM's output and clean output. }
\label{tab:Main_table}
\scalebox{1.0}{
\setlength{\tabcolsep}{1.0mm}{
\begin{tabular}{c|c|ccc|ccc|ccc}
\hline
\multirow{2}{*}{LVLM} & \multirow{2}{*}{Attack} & \multicolumn{3}{c|}{Dataset: MS-COCO} & \multicolumn{3}{c|}{Dataset: ImageNet} & \multicolumn{3}{c}{Dataset: DALL-E} \\ \cline{3-11} 
 & & Cls. & Cap. & VQA & Cls. & Cap. & VQA & Cls. & Cap. & VQA \\ \hline

\multirow{4}{*}{BLIP-2} 
 & Clean$^{Tar}$ & 0.409 & 0.436 & 0.447 & 0.415 & 0.462 & 0.473 & 0.368 & 0.425 & 0.466 \\
 & \textit{Ours}$^{Tar}$ & \textbf{0.862} & \textbf{0.833} & \textbf{0.827} & \textbf{0.831} & \textbf{0.814} & \textbf{0.860} & \textbf{0.802} & \textbf{0.841} & \textbf{0.848} \\ \cline{2-11}
 & Clean$^{Untar}$ & 1.000 & 1.000 & 1.000 & 1.000 & 1.000 & 1.000 & 1.000 & 1.000 & 1.000 \\
 & \textit{Ours}$^{Untar}$ & \textbf{0.524} & \textbf{0.601} & \textbf{0.547} & \textbf{0.543} & \textbf{0.582} & \textbf{0.556} & \textbf{0.539} & \textbf{0.594} & \textbf{0.525} \\ \hline

\multirow{4}{*}{MiniGPT-4} 
 & Clean$^{Tar}$ & 0.438 & 0.451 & 0.463 & 0.419 & 0.447 & 0.504 & 0.396 & 0.441 & 0.497 \\
 & \textit{Ours}$^{Tar}$ & \textbf{0.849} & \textbf{0.815} & \textbf{0.872} & \textbf{0.837} & \textbf{0.862} & \textbf{0.879} & \textbf{0.816} & \textbf{0.847} & \textbf{0.864} \\ \cline{2-11}
 & Clean$^{Untar}$ & 1.000 & 1.000 & 1.000 & 1.000 & 1.000 & 1.000 & 1.000 & 1.000 & 1.000 \\
 & \textit{Ours}$^{Untar}$ & \textbf{0.493} & \textbf{0.596} & \textbf{0.524} & \textbf{0.504} & \textbf{0.581} & \textbf{0.535} & \textbf{0.508} & \textbf{0.573} & \textbf{0.546} \\ \hline

\multirow{4}{*}{LLaVA-1.5} 
 & Clean$^{Tar}$ & 0.385 & 0.479 & 0.436 & 0.448 & 0.434 & 0.459 & 0.407 & 0.453 & 0.517 \\
 & \textit{Ours}$^{Tar}$ & \textbf{0.875} & \textbf{0.841} & \textbf{0.880} & \textbf{0.826} & \textbf{0.803} & \textbf{0.865} & \textbf{0.831} & \textbf{0.815} & \textbf{0.850} \\ \cline{2-11}
 & Clean$^{Untar}$ & 1.000 & 1.000 & 1.000 & 1.000 & 1.000 & 1.000 & 1.000 & 1.000 & 1.000 \\
 & \textit{Ours}$^{Untar}$ & \textbf{0.502} & \textbf{0.574} & \textbf{0.557} & \textbf{0.498} & \textbf{0.557} & \textbf{0.542} & \textbf{0.520} & \textbf{0.552} & \textbf{0.531} \\ \hline

\multirow{4}{*}{InstructBLIP} 
 & Clean$^{Tar}$ & 0.473 & 0.512 & 0.508 & 0.453 & 0.487 & 0.462 & 0.434 & 0.469 & 0.483 \\
 & \textit{Ours}$^{Tar}$ & \textbf{0.839} & \textbf{0.803} & \textbf{0.844} & \textbf{0.830} & \textbf{0.841} & \textbf{0.859} & \textbf{0.823} & \textbf{0.874} & \textbf{0.836} \\ \cline{2-11}
 & Clean$^{Untar}$ & 1.000 & 1.000 & 1.000 & 1.000 & 1.000 & 1.000 & 1.000 & 1.000 & 1.000 \\
 & \textit{Ours}$^{Untar}$ & \textbf{0.510} & \textbf{0.565} & \textbf{0.526} & \textbf{0.522} & \textbf{0.568} & \textbf{0.544} & \textbf{0.515} & \textbf{0.566} & \textbf{0.537} \\ \hline
\end{tabular}}
}
\vspace{-6pt}
\end{table}

\begin{table}[t!]
\small
\centering
\caption{Performance comparison $(\uparrow)$ with other LVLM attacks on two datasets. We set $\epsilon = 8/255$ for fair comparison.}
\label{tab:compare_other}
\setlength{\tabcolsep}{0.8mm}{
\begin{tabular}{c|ccc|ccc}
\hline
\multirow{2}{*}{Attack} & \multicolumn{3}{c|}{Dataset: MS-COCO} & \multicolumn{3}{c}{Dataset: ImageNet} \\ \cline{2-7} 
 & BLIP-2 & MiniGPT-4 & LLaVA-1.5 & BLIP-2 & MiniGPT-4 & LLaVA-1.5 \\ \hline
MF-it \cite{zhao2024evaluating} & 0.602 & 0.571 & 0.549 & 0.546 & 0.484 & 0.452 \\
MF-ii \cite{zhao2024evaluating} & 0.641 & 0.623 & 0.567 & 0.592 & 0.572 & 0.450 \\
MF-ii+it \cite{zhao2024evaluating} & 0.712 & 0.703 & 0.681 & 0.665 & 0.666 & 0.597 \\
QueryAtt \cite{chen2020hopskipjumpattack} & 0.738 & 0.726 & 0.709 & 0.689 & 0.673 & 0.658 \\
Qeba \cite{li2020qeba} & 0.754 & 0.742 & 0.768 & 0.702 & 0.696 & 0.714 \\
AT-PGD \cite{cui2024robustness} & 0.781 & 0.762 & 0.793 & 0.746 & 0.711 & 0.754 \\
AT-CW \cite{cui2024robustness} & 0.768 & 0.795 & 0.812 & 0.729 & 0.763 & 0.782 \\
Adv. Patch \cite{brown2017adversarial} & 0.698 & 0.719 & 0.732 & 0.636 & 0.658 & 0.664 \\
Anydoor \cite{lu2024test} & 0.784 & 0.791 & 0.808 & 0.742 & 0.731 & 0.759 \\ \hline
\textbf{Ours} & \textbf{0.841} & \textbf{0.845} & \textbf{0.865} & \textbf{0.835} & \textbf{0.859} & \textbf{0.831} \\ \hline
\end{tabular}
}
\vspace{-12pt}
\end{table}

\begin{table}[t!]
    \centering
     \caption{Performance ($\uparrow$) on more LVLMs on ImageNet \cite{deng2009imagenet}.}
    \setlength{\tabcolsep}{1.6mm}{
    \begin{tabular}{c|cc|cc}
    \hline
    Metric & Qwen2-VL & Intern-VL & GPT-4v & Claude-3.5 \\
    \hline
    Overall & 0.814 & 0.798 & 0.759 & 0.772 \\
    \hline
    \end{tabular}}
    \vspace{-8pt}
    \label{tab:gpt}
\end{table}

\begin{table}[t!]
    \centering
    \caption{Evaluations ($\uparrow$) on more LVLMs on ImageNet \cite{deng2009imagenet}.}
    \setlength{\tabcolsep}{0.8mm}{
    \begin{tabular}{c|ccccc}
    \hline
    LVLMs & AT-PGD & AT-CW & Adv. Patch & Anydoor & Ours \\ \hline
    Qwen2-VL & 0.698 & 0.687 & 0.659 & 0.692 & \textbf{0.814}\\
    Intern-VL & 0.683 & 0.695 & 0.664 & 0.676 & \textbf{0.798} \\
    \hline
    \end{tabular}}
    \vspace{-10pt}
    \label{tab:model}
\end{table}

\begin{table}[t!]
\centering
\small
\caption{\blue{Comparison across different attack settings.}}
\setlength{\tabcolsep}{2mm}
\begin{tabular}{c|c|c|c}
\hline
\blue{Method} & \blue{Access Level} & \blue{Attack Success ($\uparrow$)} & \blue{Queries} \\ \hline
\blue{PGD} & \blue{White-box} & \blue{0.91} & \blue{10} \\
\blue{C\&W} & \blue{White-box} & \blue{0.93} & \blue{20} \\ \hline
\blue{GrayAttack} & \blue{Gray-box} & \blue{0.85} & \blue{50} \\ \hline
\blue{NES} & \blue{Black-box (score)} & \blue{0.72} & \blue{1000+} \\
\blue{Global Noise} & \blue{Black-box (hard-label)} & \blue{0.69} & \blue{300} \\
\blue{Ours} & \blue{Black-box (hard-label)} & \blue{0.81} & \blue{150} \\ 
\hline
\end{tabular}
\vspace{-8pt}
\label{tab:setting_compare}
\end{table}

\noindent \textbf{Implementation Details.}
For each input image, the patch number $M$ is set to $49$.
We follow previous work \cite{zhao2024evaluating} to employ the CLIP model \cite{radford2021learning} to evaluate the semantic similarity in Eq.~(\ref{eq:1}). 
The optimization number $T$ for each patch is set to 100, and the additive noise number $K$ is set to 200.
As for the adversarial condition, the similarity threshold $\tau_1$ for the targeted attack is set to 0.8, and the similarity threshold $\tau_2$ for the untargeted attack is set to 0.6.
The preset maximum adversarial patch number is 4.
We impose $\epsilon = 16/255$ as the constraint for most experiments. All experiments are conducted on eight NVIDIA H100 Tensor Core GPUs.

\begin{table}[t!]
\small
\centering
\caption{Targeted attack performance $(\uparrow)$ of our \textit{HardPatch} on different LVLM models on MS-COCO dataset \cite{lin2014microsoft} with different target texts.}
\label{tab:different_target}
\scalebox{1.0}{
\setlength{\tabcolsep}{0.8mm}{
\begin{tabular}{c|c|ccc}
\hline
Target Text                  & LVLM               & Classification  & Captioning & VQA  \\ \hline
\multirow{4}{*}{I am sorry} & BLIP-2 & 0.801 & 0.792 & 0.837  \\ 
                         & MiniGPT-4 	& 0.850 & 0.834 & 0.817  \\
                         & LLaVA-1.5 	& 0.862 & 0.841 & 0.874  \\ 
                         & InstructBLIP 	& 0.834 & 0.803 & 0.825 \\ 
                         \cline{1-5}
\multirow{4}{*}{I do not know} & BLIP-2  & 0.878 & 0.775 & 0.861  \\ 
                         & MiniGPT-4 & 0.825 & 0.809 & 0.842   \\
                         & LLaVA-1.5 	& 0.857 & 0.825 & 0.853  \\ 
                         & InstructBLIP 	& 0.836 & 0.799 & 0.828 \\ 
                         \cline{1-5}
\multirow{4}{*}{I cannot answer} & BLIP-2  & 0.843 & 0.816 & 0.839  \\ 
                         & MiniGPT-4 	& 0.864 & 0.827 & 0.848  \\
                         & LLaVA-1.5 	& 0.872 & 0.824 & 0.866  \\ 
                         & InstructBLIP 	& 0.821 & 0.790 & 0.809 \\ 
                         \cline{1-5}
\multirow{4}{*}{Bomb} & BLIP-2  & 0.835 & 0.804 & 0.851  \\ 
                         & MiniGPT-4 	& 0.819 & 0.843 & 0.820 \\
                         & LLaVA-1.5 	& 0.830 & 0.798 & 0.842  \\ 
                         & InstructBLIP 	& 0.806 & 0.782 & 0.815 \\ 
\hline
\end{tabular}
}}
\vspace{-12pt}
\end{table}

\subsection{Performance on Targeted/Untargeted Setting}
To evaluate the effectiveness of the proposed \textit{HardPatch} attack, we show attack performance on different LVLM models in Table~\ref{tab:Main_table}. Here, we implement our \textit{HardPatch} in both targeted and untargeted attack settings. As for the targeted attack, we report the semantic similarities between the LVLM’s output and the attackers’ chosen label, where the larger score denotes better performance. 
As for the untargeted attack, we report the semantic similarities between the LVLM’s output and clean output, where the smaller score denotes better performance. 
From this table, we can conclude that:
(1) As for the targeted attack, the output of clean images Clean$^{Tar}$ shares low textual semantic similarity with the target text. By only querying the LVLM model, \textit{Ours}$^{Tar}$ can significantly guide the model's output to fit the target text with much higher similarity.
(2) As for the untargeted attack, \textit{Ours}$^{Untar}$ can keep the model's output away from the clean output with much smaller similarity.
We also compare our attack with previous LVLM attackers MF \cite{zhao2024evaluating} and AT \cite{cui2024robustness}, query-based attacks QueryAtt \cite{chen2020hopskipjumpattack} and Qeba \cite{li2020qeba}, and other patch-based attack methods Adv. Patch \cite{brown2017adversarial} and Anydoor \cite{lu2024test} for fair comparison in Table~\ref{tab:compare_other}, where our attack still achieves better performance. Table~\ref{tab:gpt} and Table~\ref{tab:model} also indicate that our method can attack realistic LVLM applications with high similarities. 

\blue{We further compare our method with representative attacks under different access levels in Table~\ref{tab:setting_compare}. As expected, white-box methods (e.g., PGD, C\&W) achieve the highest attack success due to full gradient access. Gray-box methods provide a trade-off between access and performance. 
Importantly, under the strictly harder hard-label black-box setting, our method achieves competitive performance and significantly outperforms existing black-box baselines. The gap between our method and gray-box approaches is relatively small, demonstrating the effectiveness of our design in exploiting model vulnerabilities even without internal access.
This result suggests that our framework successfully bridges part of the performance gap between black-box and more privileged attack settings, while maintaining practical applicability.}

To demonstrate that the effectiveness of the proposed \textit{HardPatch} method is not constrained to the specific case of the target text “unknown”, we also extend our evaluation to various other target texts in Table~\ref{tab:different_target}. The experiment includes a selection of text with varied length and usage frequency.
We can observe that our \textit{HardPatch} attack performs the best overall and in each individual task under different target text, though the similarity differs for different target prompts. 
In summary, our \textit{HardPatch} can effectively attack the LVLMs in the challenging hard-label setting.

\begin{figure*}[t!]
\centering
\includegraphics[width=0.99\textwidth]{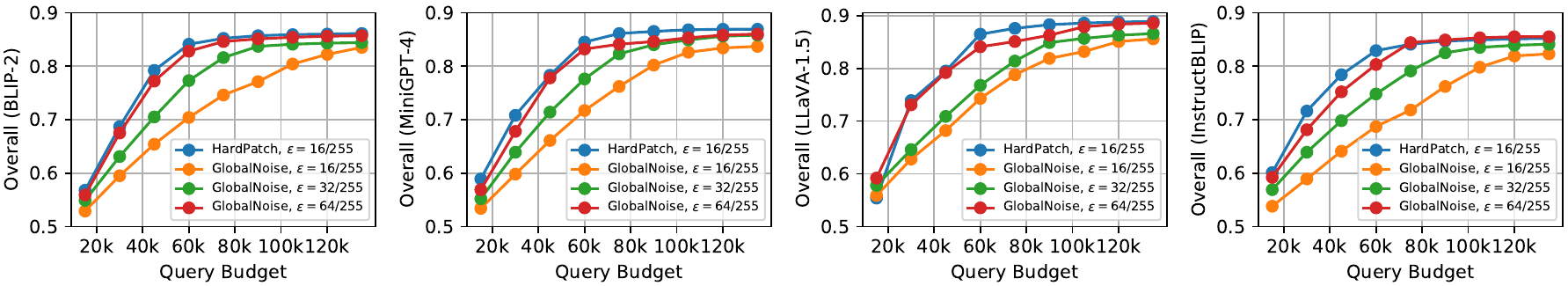}
\vspace{-10pt}
\caption{Performance comparison between our adversarial patch and the global noise on the MS-COCO dataset \cite{lin2014microsoft}.}
\vspace{-6pt}
    \label{fig:local_global}
\end{figure*} 

\begin{figure*}[t!]
\centering
\includegraphics[width=0.99\textwidth]{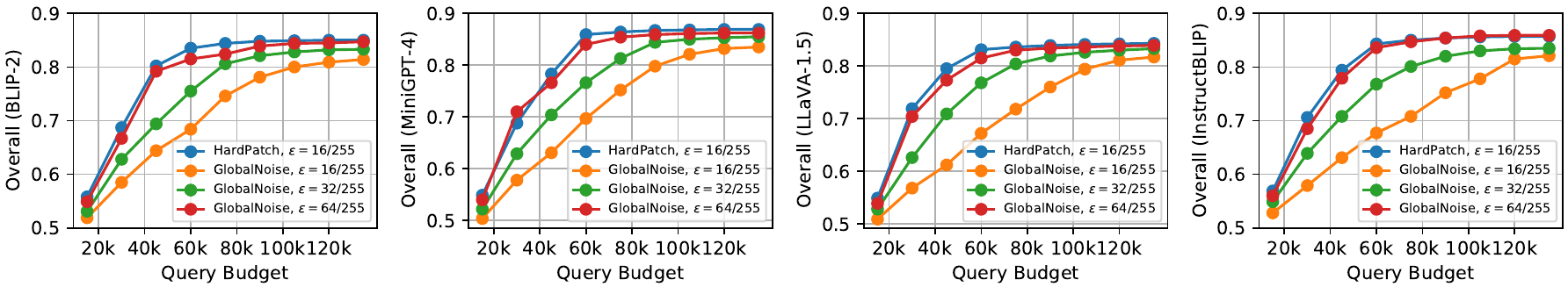}
\vspace{-10pt}
\caption{Performance comparison between our adversarial patch and the global noise on the ImageNet dataset \cite{deng2009imagenet}.}
\vspace{-6pt}
    \label{fig:more_lg1}
\end{figure*} 

\begin{figure*}[t!]
\centering
\includegraphics[width=0.99\textwidth]{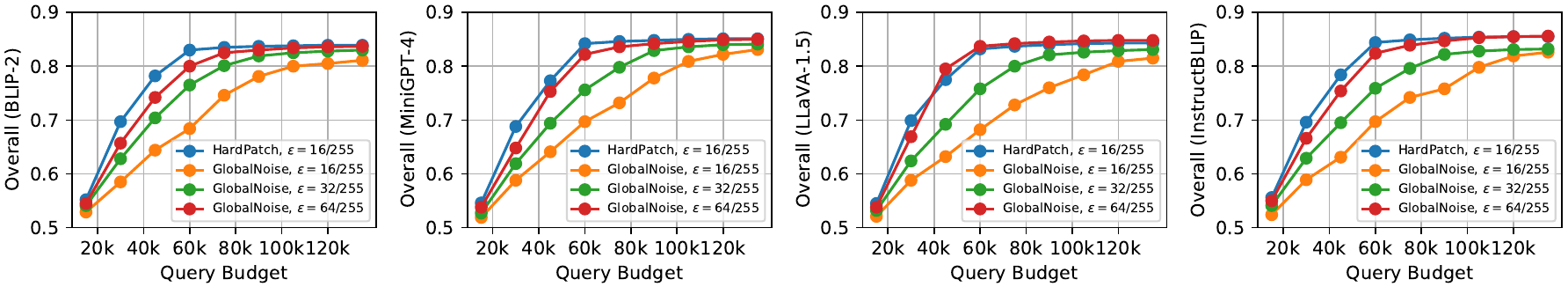}
\vspace{-10pt}
\caption{Performance comparison between our adversarial patch and the global noise on the DALL-E dataset \cite{ramesh2021zero,ramesh2022hierarchical}.}
\vspace{-6pt}
    \label{fig:more_lg2}
\end{figure*} 

\begin{figure}[t!]
\centering
\includegraphics[width=0.98\textwidth]{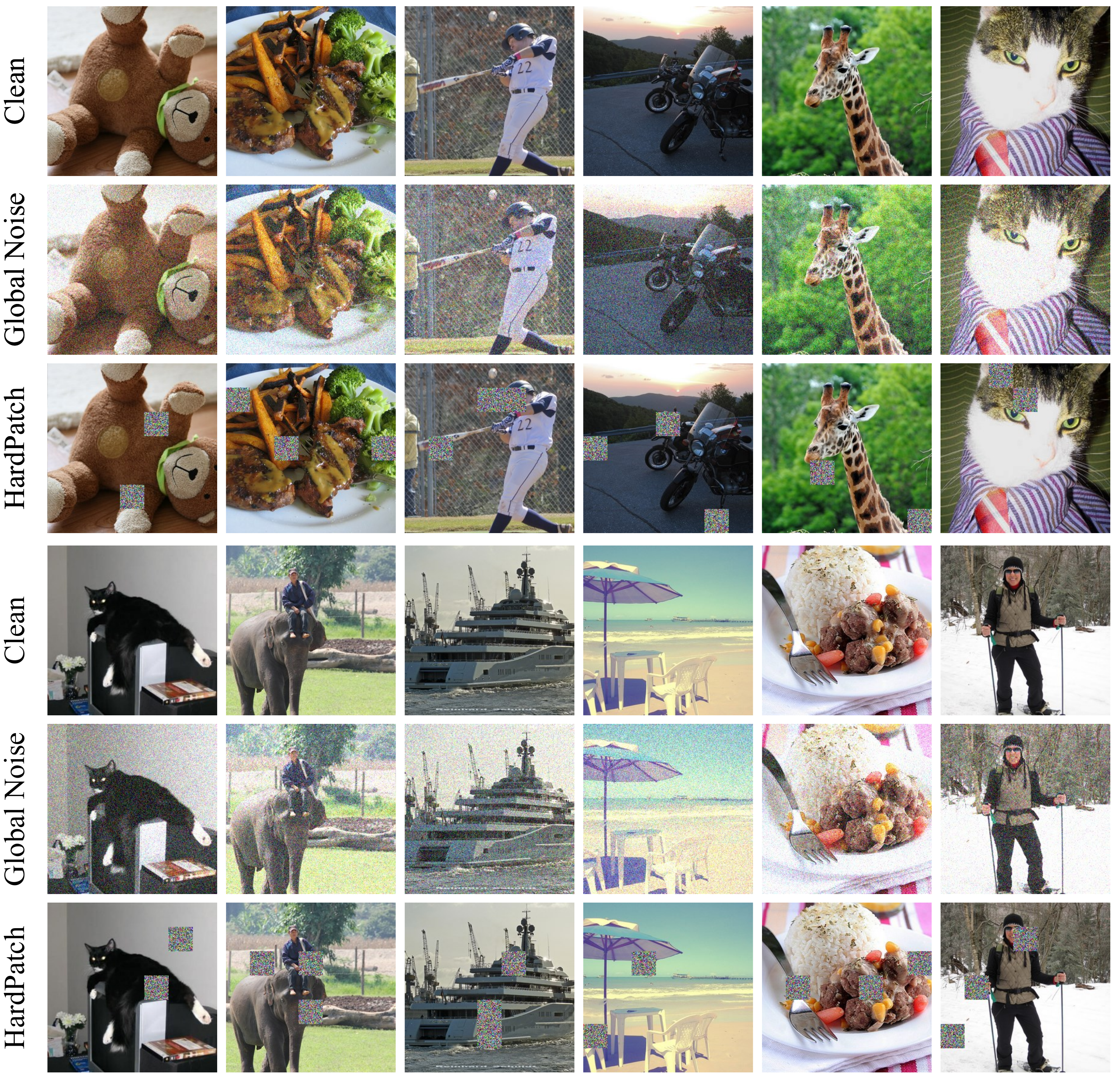}
\caption{Visualization of the adversarial examples generated by our \textit{HardPatch} and the global noise on LLaVA-1.5 \cite{liu2024visual} under the targeted attack.}
    \label{fig:locao_global_visualization}
\end{figure} 

\subsection{Adversarial Patch \textit{vs.} Global Noise?}
We provide an in-depth analysis of why we should choose the adversarial patch instead of the global noise for attacking hard-label LVLMs.
In the hard-label setting, we can not explicitly know how LVLM models comprehend and reason the input image according to the prompt.
Therefore, without understanding the vulnerability of local image regions, directly adding and optimizing global noise to all pixels of the whole image (using Monte Carlo strategy) makes it difficult to achieve good performance as its optimization/search space is too large and complicated.
Unlike this global noise, our \textit{HardPatch} attack is able to implicitly perceive the patch-wise sensitivity to the LVLM model for determining the substitution and optimization location of adversarial patches. 
We provide detailed experiments on four LVLMs in Figure~\ref{fig:local_global}, \ref{fig:more_lg1}, \ref{fig:more_lg2}. Under the same perturbation budget $\epsilon=16/255$, global noise requires much more query steps and times (about 2$\times$) for optimization, and also achieves relatively worse performance.
Although global noise with larger $\epsilon=64/255$ can achieve similar performance with our method, it significantly increases the noise size, resulting in low-quality and noticeable perturbed images.
Therefore, our adversarial patch is more imperceptible and efficient.
More experiments and visualizations are illustrated in Figure~\ref{fig:locao_global_visualization}. 
It shows that global noise is very large and noticeable, while our adversarial patch is easier to add to the images and is relatively more imperceptible.
\blue{Compared with global-noise attacks, our localized patch design significantly reduces the spatial extent of perturbations, making them less intrusive and more feasible for real-world deployment.}

\begin{figure*}[t!]
\centering
\includegraphics[width=0.99\textwidth]{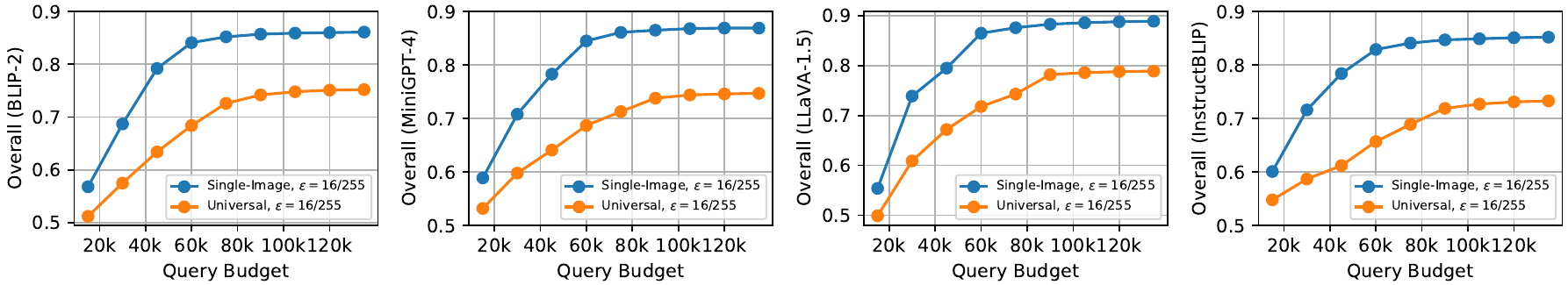}
\vspace{-10pt}
\caption{Performance comparison of our \textit{HardPatch} in single-image and universal attack settings on the MS-COCO dataset \cite{lin2014microsoft}.}
\vspace{-6pt}
    \label{fig:universal}
\end{figure*} 

\begin{figure*}[t!]
\centering
\includegraphics[width=0.99\textwidth]{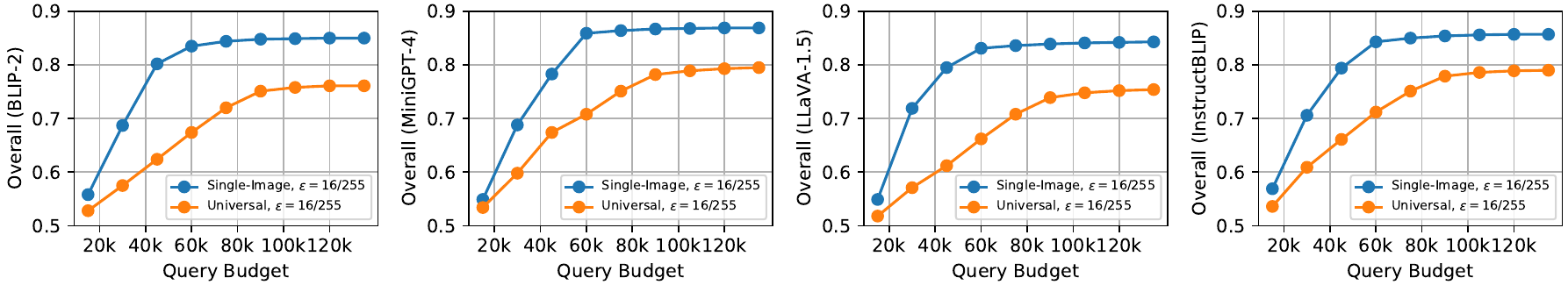}
\vspace{-10pt}
\caption{Performance comparison of our \textit{HardPatch} in single-image and universal attack settings on the ImageNet dataset \cite{deng2009imagenet}.}
\vspace{-6pt}
    \label{fig:universal_more1}
\end{figure*} 

\begin{figure*}[t!]
\centering
\includegraphics[width=0.99\textwidth]{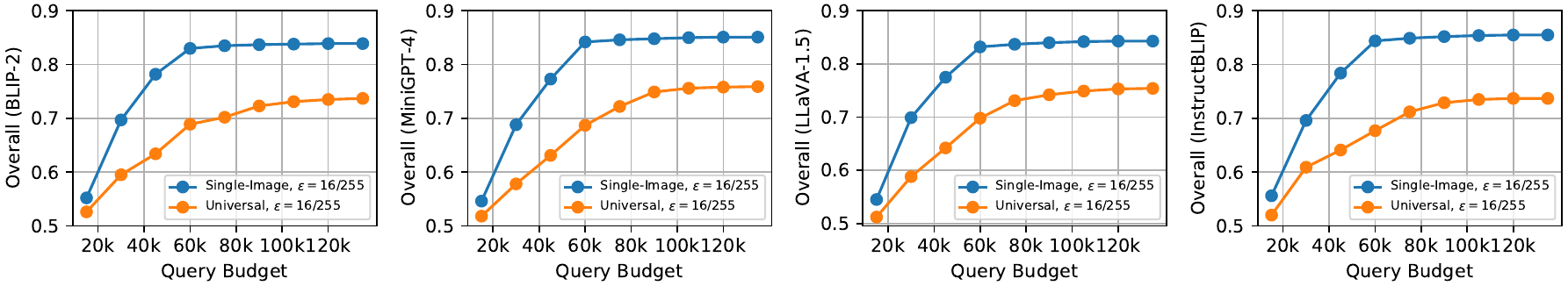}
\vspace{-10pt}
\caption{Performance comparison of our \textit{HardPatch} in single-image and universal attack settings on the DALL-E dataset \cite{ramesh2021zero,ramesh2022hierarchical}.}
\vspace{-6pt}
    \label{fig:universal_more2}
\end{figure*}

\subsection{Extending \textit{HardPatch} to Universal Attack Setting}
\begin{figure}[t!]
\centering
\includegraphics[width=0.98\textwidth]{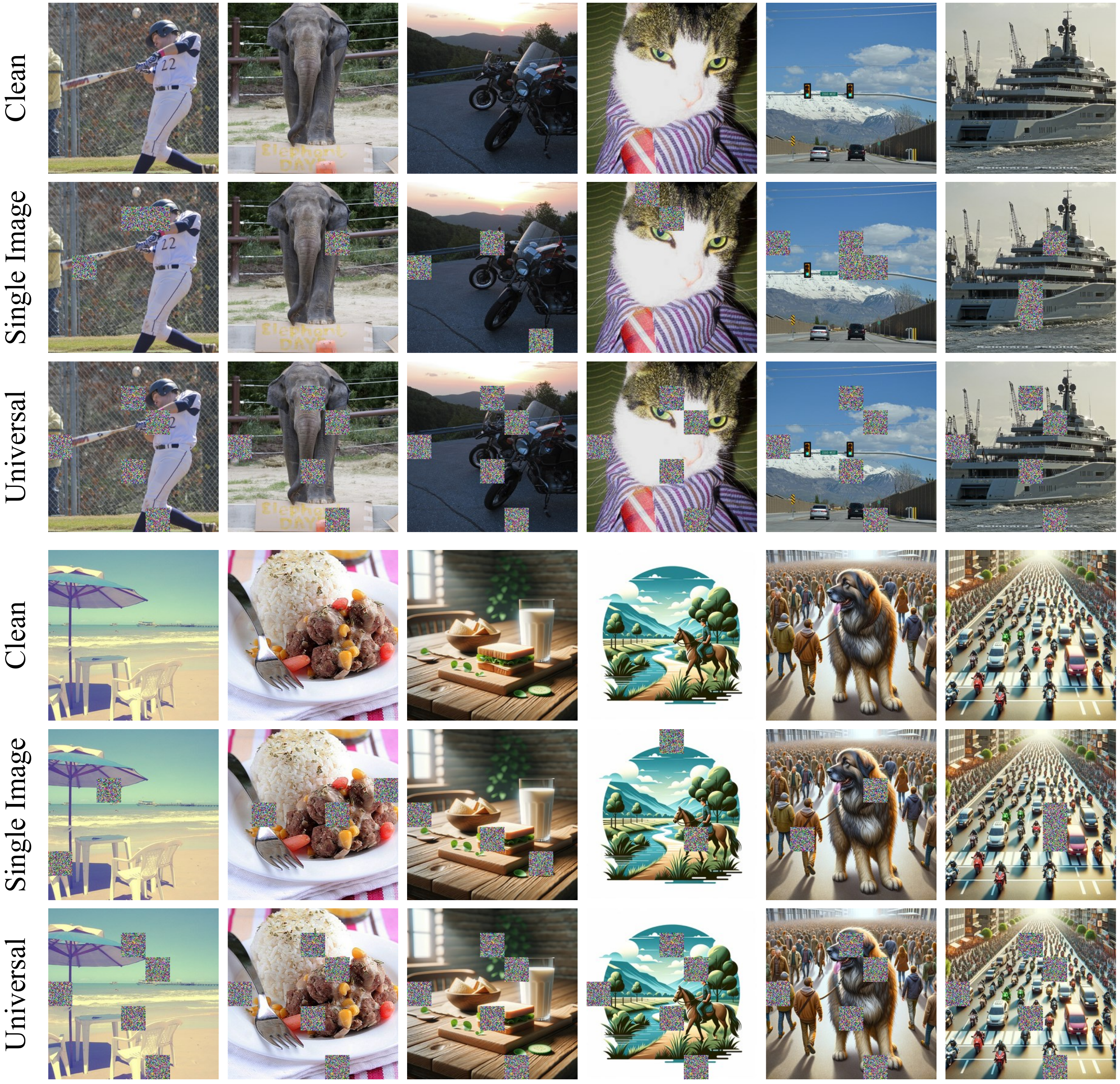}
\caption{Visualization of the adversarial examples generated with LLaVA-1.5 \cite{liu2024visual} in both single-image attack and universal attack settings.}
    \label{fig:universality_visualization}
\end{figure}

In all our experiments, we implement our proposed \textit{HardPatch} method in a single-image attack setting, where the perturbed patches vary among different image-text inputs.
Further, we can also extend our \textit{HardPatch} attack into a universal attack setting, where the patches are the same among all image-text input.
Specifically, we follow the traditional universal setting \cite{moosavi2017universal} to optimize vulnerable patches.
In particular, we first assess the sensitivities of all patches based on their averaged impacts on the whole test set. Then, we jointly optimize the patches in their descending order to attack all image-prompt inputs. As shown in Figure~\ref{fig:universal}, \ref{fig:universal_more1}, \ref{fig:universal_more2}, we can conclude that:
(1) In the same perturbation budget, the universal attack setting is much more difficult to achieve since different images share diverse sensitive regions in different locations to the LVLM model. Therefore, it requires more querying steps and achieves lower final performance in the targeted attack setting.
(2) Instead, the single-image attack is more flexible and can straightforwardly perturb the most vulnerable patches in each image. Therefore, it is more efficient and can achieve better attack performance.
Visualization comparisons are further shown in Figure~\ref{fig:universality_visualization}, where
the universal attack setting is much more difficult to achieve since different images share diverse sensitive regions in different locations to the LVLM model, requiring a larger number of adversarial patches.

\begin{table}[t!]
\small
\centering
\caption{Targeted attack performance $(\uparrow)$ of our \textit{HardPatch} on MS-COCO dataset \cite{lin2014microsoft} with different maximum adversarial patch number.}
\label{tab:different_patch_number}
\scalebox{1.0}{
\setlength{\tabcolsep}{0.8mm}{
\begin{tabular}{c|c|ccc}
\hline
Max Number                  & LVLM                & Classification  & Captioning & VQA  \\ \hline
\multirow{4}{*}{Number$=1$} & BLIP-2 & 0.678 & 0.642 & 0.651  \\ 
                         & MiniGPT-4  & 0.649 & 0.665 & 0.670 \\
                         & LLaVA-1.5  & 0.626 & 0.634 & 0.668 	\\ 
                         & InstructBLIP 	& 0.681 & 0.652 & 0.645  \\ 
                         \cline{1-5}
\multirow{4}{*}{Number$=2$} & BLIP-2  & 0.749 & 0.726 & 0.768  \\ 
                         & MiniGPT-4 & 0.761 & 0.704 & 0.753  \\
                         & LLaVA-1.5  & 0.757 & 0.725 & 0.752 \\ 
                         & InstructBLIP & 0.772 & 0.730 & 0.746 \\ 
                         \cline{1-5}
\multirow{4}{*}{Number$=3$} & BLIP-2  & 0.822 & 0.804 & 0.800  \\ 
                         & MiniGPT-4  & 0.815 & 0.793 & 0.828  \\
                         & LLaVA-1.5  & 0.861 & 0.807 & 0.836 	\\ 
                         & InstructBLIP & 0.810 & 0.779 & 0.814 \\ 
                         \cline{1-5}
\multirow{4}{*}{Number$=4$} & BLIP-2  & 0.862 & 0.833 & 0.827  \\ 
                         & MiniGPT-4  & 0.849 & 0.815 & 0.872  \\
                         & LLaVA-1.5  & 0.875 & 0.841 & 0.880 	\\ 
                         & InstructBLIP & 0.839 & 0.803 & 0.844 \\ 
\hline
\end{tabular}
}}
\vspace{-12pt}
\end{table}

\subsection{Further Analysis}
\noindent \textbf{The Influence of the Maximum Number of Adversarial Patches.}
The number of adversarial patches is related to the imperceptibility. 
Since more adversarial patches will mask most image contents and lead to noticeable noise (which is also not meaningful), in our attack algorithm, we preset the maximum number of adversarial patches to a fixed number of 4. That means, only $\{1,2,3,4\}$ adversarial patches may be added to the image.
To investigate the influence of the maximum number of adversarial patches on the adversarial conditions, we conduct corresponding experiments in Table~\ref{tab:different_patch_number}.
We can conclude that:
(1) Only one adversarial patch is not enough to mask and perturb most images' semantics, resulting in relatively lower attack performance.
(2) More adversarial patches can better fool the LVLM model with more vulnerable visual contents.
(3) Four adversarial patches are enough to achieve great attack performance. 
Of course, the adversarial patch number larger than 4 can further boost the attack performance.
However, considering more adversarial patches cost more resources and time, we preset the adversarial patch number to 4 in all our experiments.

\begin{table}[t!]
\small
\centering
\caption{Targeted attack performance $(\uparrow)$ of our \textit{HardPatch} on MS-COCO dataset \cite{lin2014microsoft} with different image split. The maximum adversarial patch number is set to 4.}
\label{tab:different_image_split}
\scalebox{1.0}{
\setlength{\tabcolsep}{0.8mm}{
\begin{tabular}{c|c|ccc}
\hline
Image Split $M$                  & LVLM               & Classification  & Captioning & VQA  \\ \hline
\multirow{4}{*}{Split to $5 \times 5$} & BLIP-2  & 0.881 & 0.842 & 0.839  \\ 
                         & MiniGPT-4  & 0.875 & 0.830 & 0.863  \\
                         & LLaVA-1.5  & 0.874 & 0.836 & 0.872	\\ 
                         & InstructBLIP & 0.868 & 0.824 & 0.850\\ 
                         \cline{1-5}
\multirow{4}{*}{Split to $7 \times 7$} & BLIP-2  & 0.862 & 0.833 & 0.827  \\ 
                         & MiniGPT-4  & 0.849 & 0.815 & 0.872 \\
                         & LLaVA-1.5  & 0.875 & 0.841 & 0.880 	\\ 
                         & InstructBLIP & 0.839 & 0.803 & 0.844 \\  \cline{1-5}
\multirow{4}{*}{Split to $9 \times 9$} & BLIP-2  & 0.849 & 0.821 & 0.816  \\ 
                         & MiniGPT-4 & 0.834 & 0.801 & 0.852  \\
                         & LLaVA-1.5  & 0.861 & 0.829 & 0.870 	\\ 
                         & InstructBLIP & 0.827 & 0.789 & 0.833 \\
\hline
\end{tabular}
}}
\vspace{-6pt}
\end{table}

\begin{table}[t!]
\small
\centering
\caption{Targeted attack performance $(\uparrow)$ of our \textit{HardPatch} on different patch orders on MS-COCO \cite{lin2014microsoft} dataset. The maximum adversarial patch number is set to 4.}
\label{tab:ablation_1}
\scalebox{1.0}{
\setlength{\tabcolsep}{0.6mm}{
\begin{tabular}{c|c|ccc}
\hline
Patch Order                  & LVLM                & Classification  & Captioning & VQA \\ \hline
\multirow{4}{*}{Random Order} & BLIP-2  & 0.714 & 0.697 & 0.680\\ 
                         & MiniGPT-4  & 0.696  & 0.672 & 0.733 \\
                         & LLaVA-1.5  & 0.729 & 0.703 & 0.737  	\\ 
                         & InstructBLIP & 0.688 & 0.675 & 0.699 \\ 
                         \cline{1-5}
\multirow{4}{*}{Descending Order} & BLIP-2 & 0.862 & 0.833 & 0.827 \\ 
                         & MiniGPT-4  & 0.849 & 0.815 & 0.872  \\
                         & LLaVA-1.5  & 0.875 & 0.841 & 0.880 	\\ 
                         & InstructBLIP & 0.839 & 0.803 & 0.844\\  \hline
\end{tabular}
}}
\vspace{-12pt}
\end{table}

\noindent \textbf{Performance of Attack with Different Image Split.}
We also investigate the impact of different settings of image split. In all our experiments, we split each image into $7 \times 7$ patches. As shown in Table~\ref{tab:different_image_split}, we conduct experiments on the image split of $5 \times 5$ and $9 \times 9$, respectively. We can conclude that:
Different image splits of the same maximum adversarial patch number share similar attack performances. Since patches in $5 \times 5$ split have more perturbed pixels, it is easier to achieve the attack. Instead, patches in $9 \times 9$ split have fewer perturbed pixels, thus achieving a lower performance. Therefore, we set the split of each image into $7 \times 7$ patches in all our experiments.

\noindent \textbf{Effectiveness of the Replacement Order Determination.}
To demonstrate the effectiveness of our proposed module of Replacement Order Determination, we conduct an ablation study in Table~\ref{tab:ablation_1} where we change our LVLM-sensitive replacement order into a random version. From this table, we can conclude that:
(1) Random order may select LVLM's insensitive patches, resulting in more difficult patch optimization for achieving attack.
(2) Our Replacement Order Determination can assess the vulnerability of each patch, and provide a descending order for easily achieving attack.
Therefore, the proposed Replacement Order Determination module can help efficiently and effectively find the global optimal patches for perturbation.

\begin{table}[t!]
\small
\centering
\caption{Targeted attack performance $(\uparrow)$ of our \textit{HardPatch} against defense strategy of RandomRotation on MS-COCO \cite{lin2014microsoft} dataset.}
\label{tab:defense}
\setlength{\tabcolsep}{0.6mm}{
\begin{tabular}{c|c|cccc}
\hline
Defense                 & LVLM                & Classification  & Captioning & VQA  \\ \hline
\multirow{4}{*}{With Defense} & BLIP-2 & 0.828 & 0.779  & 0.781   \\ 
                         & MiniGPT-4  & 0.797 & 0.762 & 0.803 \\
                         & LLaVA-1.5  & 0.815 & 0.784 & 0.810	\\ 
                         & InstructBLIP & 0.783 & 0.756 & 0.772\\ 
                         \cline{1-5}
\multirow{4}{*}{Without Defense} & BLIP-2  & 0.862 & 0.833 & 0.827  \\ 
                         & MiniGPT-4  & 0.849 & 0.815 & 0.872  \\
                         & LLaVA-1.5  & 0.875 & 0.841 & 0.880 	\\ 
                         & InstructBLIP & 0.839 & 0.803 & 0.844 \\  \hline
\end{tabular}}
\vspace{-6pt}
\end{table}

\begin{table}[t!]
    \small
    \centering
    \caption{Averaged Attack performance ($\uparrow$) on three tasks against more complicated defenses on ImageNet \cite{deng2009imagenet} dataset.}
    \setlength{\tabcolsep}{1.6mm}{
    \begin{tabular}{c|ccc}
    \hline
    Defense & BLIP-2 & MiniGPT-4 & LLaVA-1.5 \\
    \hline
    None & 0.835 & 0.859 & 0.831 \\
    Simple Smoothing & 0.748 & 0.756 & 0.772\\
    JPEG Compression & 0.762 & 0.748 & 0.756 \\
    Patch Diffusion & 0.725 & 0.713 & 0.709 \\
    Color Distortion & 0.777 & 0.794 & 0.770 \\
    \hline
    \end{tabular}}
    \vspace{-6pt}
    \label{tab:defense_add}
\end{table}

\noindent \textbf{Robustness to Defense Strategy.}
To evaluate the robustness of our proposed \textit{HardPatch} attack, we follow previous work \cite{luo2024image} to exploit widely used RandomRotation as the defense strategy to defend our generated adversarial examples on four LVLM models.
As shown in Table~\ref{tab:defense}, our \textit{HardPatch} just achieves slightly lower performance on the RandomRotation defense, validating that our attack is robust enough against the potential defense strategy.
We also implement more complicated defense strategies in realistic scenarios to investigate the robustness of our proposed attack. As shown in Table~\ref{tab:defense_add}, our attack is also robust to them. 

\noindent \textbf{Ablation on Similarity Calculator.}
As in Table~\ref{tab:text}, we conduct ablations on different similarity calculators (output text encoders) to measure the semantics of LVLM’s output. We find that our attack is not sensitive to it.

\noindent \textbf{Efficiency Analysis.}
As in Table~\ref{tab:efficiency}, our attack achieves competitive efficiency compared with other methods.
\blue{We further analyze the computational cost of our method in Table~\ref{tab:cost}. Compared to a single forward pass, our method incurs additional overhead due to the hard-label black-box setting. In total, our method typically requires around 100--300 forward passes.
Compared with existing black-box attacks, our method demonstrates improved efficiency. Score-based approaches such as NES or ZOO require hundreds to thousands of queries due to high-dimensional gradient estimation, while global-noise-based hard-label attacks often rely on extensive search over the entire image. In contrast, our method significantly reduces the search space via patch sensitivity estimation, leading to fewer required queries.
Although white-box attacks (e.g., PGD) may appear computationally cheaper in terms of forward passes, they rely on backward propagation and full access to model gradients, which are not available in practical LVLM scenarios. Therefore, under realistic API-based settings, our method achieves a favorable trade-off between attack effectiveness and computational cost.}


\begin{table}[t!]
    \small
    \centering
    \caption{Ablations ($\uparrow$) on output text encoders on ImageNet \cite{deng2009imagenet}.}
    \setlength{\tabcolsep}{1.6mm}{
    \begin{tabular}{c|ccc}
    \hline
    Output Text Encoder & BLIP-2 & MiniGPT-4 & LLaVA-1.5 \\
    \hline
    CLIP & 0.835 & 0.859 & 0.831 \\
    SentenceTransformer & 0.842 & 0.861 & 0.837 \\
    \hline
    \end{tabular}}
    \vspace{-6pt}
    \label{tab:text}
\end{table}

\begin{table}[t!]
\small
\centering
\caption{Analysis on the method efficiency of our \textit{HardPatch}.}
\label{tab:efficiency}
\setlength{\tabcolsep}{0.8mm}{
    \begin{tabular}{c|ccccc}
    \hline
    ~ & MF \cite{zhao2024evaluating} & AT \cite{cui2024robustness} & Adv. Patch \cite{brown2017adversarial} & Anydoor \cite{lu2024test} & Ours \\
    \hline
    Time & 12.2min & 1.8min & 3.5min & 2.6min & 2.4min \\
    \hline
    \end{tabular}}
\vspace{-6pt}
\end{table}

\begin{table}[t!]
\centering
\caption{\blue{Computational cost comparison. Time is measured on a single GPU with batch size 1 and averaged over samples.}}
\small
\setlength{\tabcolsep}{2mm}
\begin{tabular}{c|c|c|c|c}
\hline
\blue{Method} & \blue{Access Type} & \blue{Forward} & \blue{Backward} & \blue{Time (s)} \\ \hline
\blue{Single Inference} & \blue{--} & \blue{1} & \blue{0} & \blue{$\sim$0.1} \\
\blue{PGD (10 iters)} & \blue{White-box} & \blue{10} & \blue{10} & \blue{$\sim$1--2} \\
\blue{NES / ZOO} & \blue{Score-based} & \blue{500--2000} & \blue{0} & \blue{$\sim$10--60} \\
\blue{Global Noise (Hard-label)} & \blue{Label-only} & \blue{200--500} & \blue{0} & \blue{$\sim$5--20} \\
\blue{Ours} & \blue{Label-only} & \blue{100--300} & \blue{0} & \blue{$\sim$3--10} \\
\hline
\end{tabular}
\vspace{-8pt}
\label{tab:cost}
\end{table}

\noindent \textbf{Investigation on Object Occlusion.}
To investigate whether the method may be leveraging occlusion rather than truly adversarial features, 
we implement masked-only controls to evaluate the attack performance without perturbation. As shown in Table~\ref{tab:occlusion}, solely masking patch cannot achieve adversarial conditions in both targeted ($\uparrow$) and untargeted ($\downarrow$) settings, demonstrating our attack leverages truly adversarial features.

\begin{table}[t!]
    \small
    \centering
    \caption{Control experiments on masked patch on MSCOCO.}
    \setlength{\tabcolsep}{0.4mm}{
    \begin{tabular}{cc|cccccc}
    \hline
    \multirow{2}*{Mask} &\multirow{2}*{Perturbation} & \multicolumn{2}{c}{BLIP-2} & \multicolumn{2}{c}{MiniGPT-4} & \multicolumn{2}{c}{LLaVA-1.5}  \\
     & & Targeted & Untargeted & Targeted & Untargeted & Targeted & Untargeted  \\ \hline
    $\checkmark$ & $\times$ & 0.463 & 0.928 & 0.478 & 0.909 & 0.463 & 0.916 \\
    $\checkmark$ & $\checkmark$ & \textbf{0.841} & \textbf{0.557} & \textbf{0.845} & \textbf{0.538} & \textbf{0.865} & \textbf{0.544}\\
    \hline
    \end{tabular}}
    \vspace{-6pt}
    \label{tab:occlusion}
\end{table}

\begin{table}[t!]
    \small
    \centering
    \caption{\blue{Evaluations ($\uparrow$) on more LVLMs on ImageNet.}}
    \setlength{\tabcolsep}{1.4mm}{
    \begin{tabular}{c|ccccc}
    \hline
    \blue{LVLMs} & \blue{AT-PGD} & \blue{AT-CW} & \blue{Adv. Patch} & \blue{Anydoor} & \blue{Ours} \\ \hline
    \blue{Qwen2-VL} & \blue{0.698} & \blue{0.687} & \blue{0.659} & \blue{0.692} & \blue{\textbf{0.814}}\\
    \blue{Intern-VL} & \blue{0.683} & \blue{0.695} & \blue{0.664} & \blue{0.676} & \blue{\textbf{0.798}} \\
    \hline
    \end{tabular}}
    \vspace{-6pt}
    \label{tab:model_more}
\end{table}

\noindent \blue{\textbf{Generalization on Different LVLM Models.}}
\blue{To evaluate the generalizability of our method beyond CLIP-based architectures, we additionally include Qwen2-VL and Intern-VL, which adopt different visual encoders and multimodal fusion strategies.
As shown in Table~\ref{tab:model_more}, our method consistently achieves the best performance.
These results suggest that our attack targets a general vulnerability in the vision-to-language generation process, rather than specific properties of CLIP-based visual encoders.}

\begin{table}[t!]
    \small
    \centering
    \caption{\blue{Evaluations ($\uparrow$) on OOD data on BLIP-2 model.}}
    \setlength{\tabcolsep}{1.4mm}{
    \begin{tabular}{c|cc}
    \hline
    & \blue{ImageNet to MSCOCO} & \blue{MSCOCO to ImageNet} \\ \hline
    \blue{OOD Image} & \blue{0.716} & \blue{0.738}\\
    \blue{OOD Prompt} & \blue{0.779} & \blue{0.781}\\
    \blue{OOD Image\&Prompt} & \blue{0.685} & \blue{0.679}\\
    \hline
    \end{tabular}}
    \vspace{-6pt}
    \label{tab:ood}
\end{table}

\noindent \blue{\textbf{Generalization to Out-of-Distribution Data.}}
\blue{We confirm that our proposed HardPatch attack demonstrates strong generalization ability under OOD settings. Specifically, we conduct cross-dataset and cross-prompt evaluations, where the attack is constructed on one dataset (or prompt distribution) and evaluated on another. As shown in Table~\ref{tab:ood}, our method maintains strong performance across different OOD scenarios. These results indicate that our attack is not overfitted to specific datasets or prompts, but instead captures more general vulnerabilities of LVLMs.}

\begin{figure*}[t!]
\centering
\includegraphics[width=0.96\textwidth]{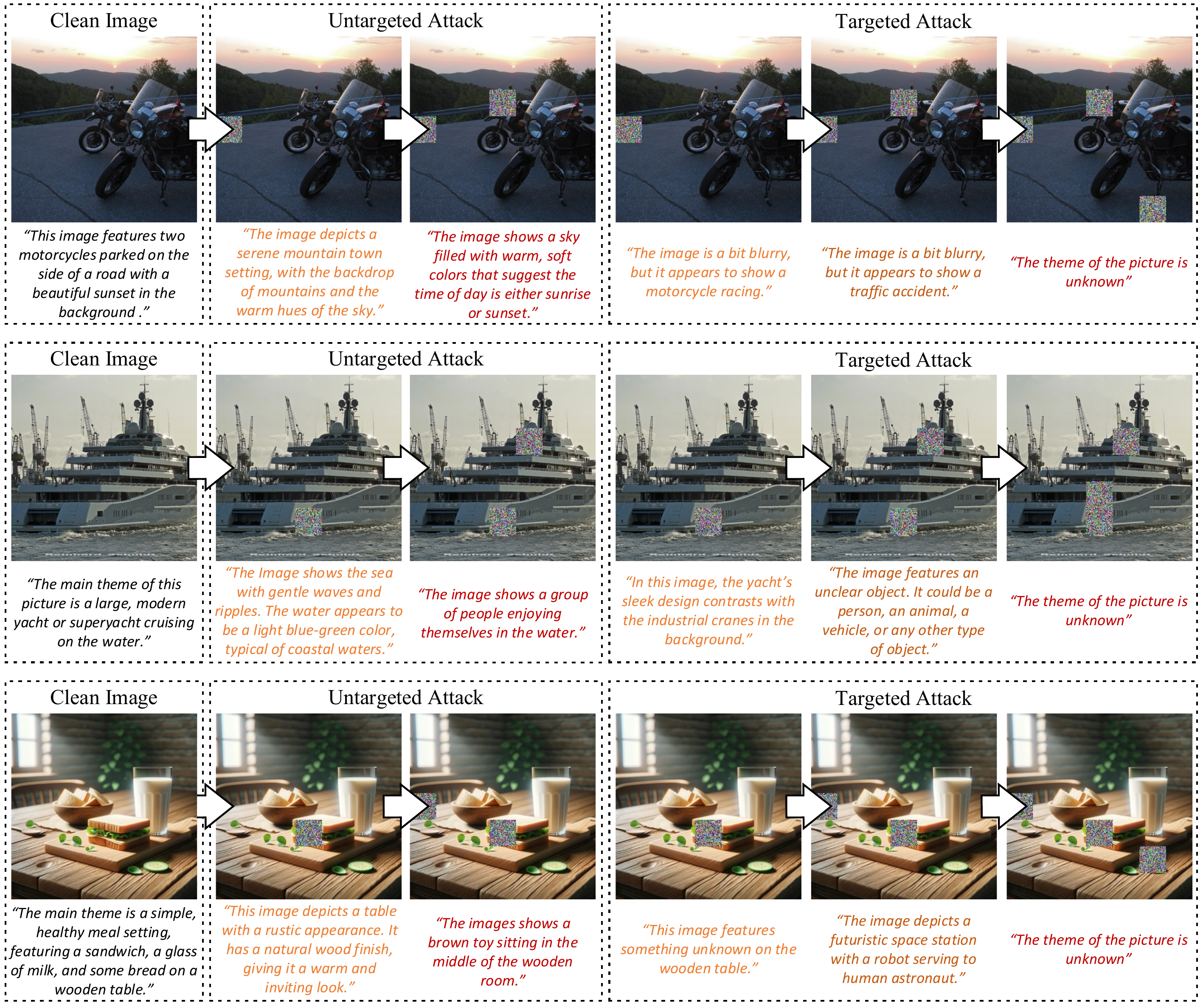}
\vspace{-10pt}
\caption{Visualizations on untargeted/targeted adversarial samples and corresponding output for the input prompt ``\textit{Convey the main theme of this picture succinctly}" on LLaVA-1.5 \cite{liu2024visual}.}
\vspace{-6pt}
    \label{fig:vis}
\end{figure*} 

\noindent \textbf{Visualizations.}
As shown in Figure~\ref{fig:vis}, we provide visualizations of the step-by-step adversarial examples and corresponding textual output of both untargeted and targeted attacks. We can conclude that the proposed \textit{HardPatch} is effective in fooling the LVLM model by dynamically changing the semantics of original images via adversarial patches.

\begin{figure}[t!]
\centering
\includegraphics[width=0.98\textwidth]{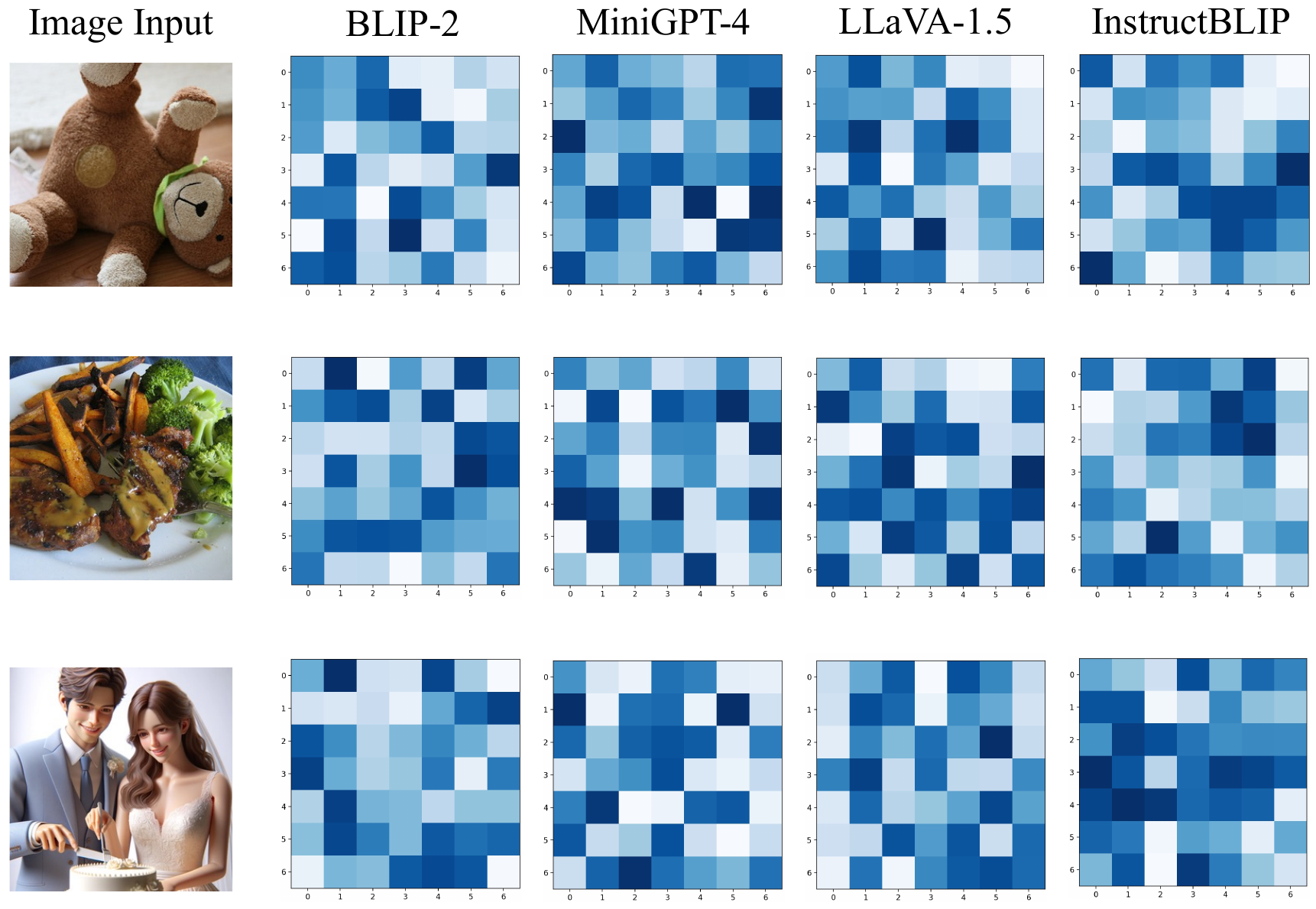}
\caption{Visualizations on the sensitivity score for each patch.}
    \label{fig:heatmap}
\end{figure} 

\subsection{Visualization on the Vulnerability of Different Patches}
At last, we visualize the sensitive scores of different patches of the same images to the LVLM model as shown in Figure~\ref{fig:heatmap}.
Here, the image is divided into $7 \times 7$ patches, and the sensitive score of each patch is measured by the semantic changes between the original output and the output of masking the corresponding patch. The heatmap of each image is computed by further using a softmax function on the scores of whole patches.
From this figure, we can conclude that:
(1) Different LVLM models have different attentions on different patches of the same image.
(2) Masking patches provide a promising way to measure the vulnerability of the LVLM models to the local regions of input images.
Based on the sensitivity scores of different patches, researchers can design specific local perturbations for attacking the LVLM models.






\subsection{Discussion of Limitations}



While our method demonstrates effectiveness across several state-of-the-art LVLMs (\textit{e.g.}, BLIP-2, LLaVA-1.5), certain limitations remain. First, our current attack is primarily evaluated in the digital domain and does not fully account for physical-world constraints, such as sensor noise, lighting variations, or camera angles, which may affect the efficacy of adversarial patches. Second, the method relies on iterative model queries to estimate gradients; in environments with strict query limits or high costs for commercial APIs, the efficiency of the attack requires further optimization. Finally, while we address general attack settings, crafting universal patches that maintain high attack success rates and imperceptibility across highly diverse datasets remains a challenging task.
\blue{While patch-based perturbations may appear more localized compared to pixel-level noise, in the hard-label setting they offer a better trade-off between attack effectiveness and perceptual quality. Compared to global perturbations, our method reduces the affected area and produces more practical adversarial patterns.}


\section{Conclusion}
In this paper, we raise a practical and challenging question, \textit{i.e.}, can visual adversarial patches fool hard-label LVLM models? In particular, we propose the first hard-label adversarial attack method called \textit{HardPatch} against LVLM models by solely querying the input/output of LVLMs. 
We start by uniformly splitting each image into multiple patches and assessing the vulnerability of LVLMs to different local patches, and then develop a patch substitution and optimization strategy to perturb the most sensitive patches with gradient estimation. 
Our empirical findings reveal that LVLMs may lose their way when appropriate patches are perturbed.
Experiments on a suite of LVLM models and datasets demonstrate the effectiveness of the proposed \textit{HardPatch} attack in the hard-label setting.
\blue{Future work will explore improving the perceptual quality of adversarial patches, such as integrating texture-aware or camouflage-based designs.}


\bibliographystyle{elsarticle-num} 
\bibliography{reference.bib}

\end{document}